\documentclass[journal]{IEEEtran}

\usepackage{graphicx}
\usepackage{cite}
\usepackage{psfrag}
\usepackage{amsmath,amssymb} 
\usepackage{color}
\usepackage{subfig}
\usepackage[lined,boxruled,linesnumberedhidden]{algorithm2e}
\usepackage{epsfig}
\graphicspath{{figure/}}

\def \R {\mathbb R}

\newcommand{\argmin}[1]{\underset{#1}{\operatorname{arg}\,\operatorname{min}}\;}
\newcommand{\argmax}[1]{\underset{#1}{\operatorname{arg}\,\operatorname{max}}\;}

\begin{document}

\title{Synthesis-based Robust Low Resolution Face Recognition}

\author{Sumit Shekhar,~\IEEEmembership{Student Member,~IEEE,}
        Vishal~M.~Patel,~\IEEEmembership{Member,~IEEE,}
        and~Rama~Chellappa,~\IEEEmembership{Fellow,~IEEE}

\thanks{Sumit Shekhar, Vishal~M.~Patel and R.~Chellappa are with the Department of Electrical and Computer Engineering and the Center for Automation Research, UMIACS, University of Maryland, College Park,
MD 20742 USA {(e-mail: \{sshekha, pvishalm, rama \}@umiacs.umd.edu})}.
}

\markboth{IEEE Transactions on Information Forensics and Security,~Vol.~X, No.~X, Month~20XX}{Synthesis-based Robust Low Resolution Face Recognition}

\maketitle

\begin{abstract}
 Recognition of low resolution face images is a challenging problem in many practical face recognition systems. Methods have been proposed in the face recognition literature for the problem which assume that the probe is low resolution, but a high resolution gallery is available for recognition. These attempts have been aimed at modifying the probe image such that the resultant image provides better discrimination. We, however, formulate the problem differently by leveraging the information available in the high resolution gallery image and proposing a generative approach for classifying the probe image. An important feature of our algorithm is that it can handle resolution change along with illumination variations. Furthermore, we also kernelize the algorithm to handle non-linearity in data and propose a joint sparse coding technique for robust recognition at low resolutions. The effectiveness of the proposed method is demonstrated using standard datasets and a challenging outdoor face dataset. It is shown that our method is efficient and can perform significantly better than many competitive low resolution face recognition algorithms.
\end{abstract}


\section{Introduction}
Face recognition (FR) has been an active field of research in biometrics for over two decades \cite{ChelSur}.  Current methods work well when the test images are captured under controlled conditions.  However, quite often the performance of most algorithms degrades significantly when they are applied to the images taken under uncontrolled conditions where there is no control over pose, illumination, expressions and resolution of the face image.  Image resolution is an important parameter in many practical scenarios such as surveillance where high resolution cameras are not deployed due to cost and data storage constraints and further, there is no control over the distance of human from the camera. 



Many methods have been proposed in the vision literature that can deal with this resolution problem in FR.  Most of these methods are based on application of super-resolution (SR) technique to increase the resolution of images so that the recovered higher-resolution (HR) images can be used for recognition.  One of the major drawbacks of applying SR techniques is that there is a possibility that recovered HR images may contain some serious artifacts.  This is often the case when the resolution of the image is very low.  As a result, these recovered images may not look like the images of the same person and the recognition performance may degrade significantly.

In practical scenarios, the resolution change is also coupled with other parameters such as pose change, illumination variations and expression.  Algorithms specifically designed to deal with LR images quite often fail in dealing with these variations.  Hence, it is essential to include these parameters while designing a robust method for low-resolution FR.  To this end, in this paper, we present a generative approach to low-resolution FR that is also robust to illumination variations based on learning class specific dictionaries.  One of the major advantages of using generative approaches is that they are known to have reduced sensitivity to noise than the discriminative approaches \cite{ChelSur}.  Furthermore, we kernelize the learning algorithm to handle non-linearity in the data samples and introduce a bi-level sparse coding framework for robust recognition. 

\begin{figure}[htp!]
\centering
\resizebox{3.5in}{!}{\includegraphics{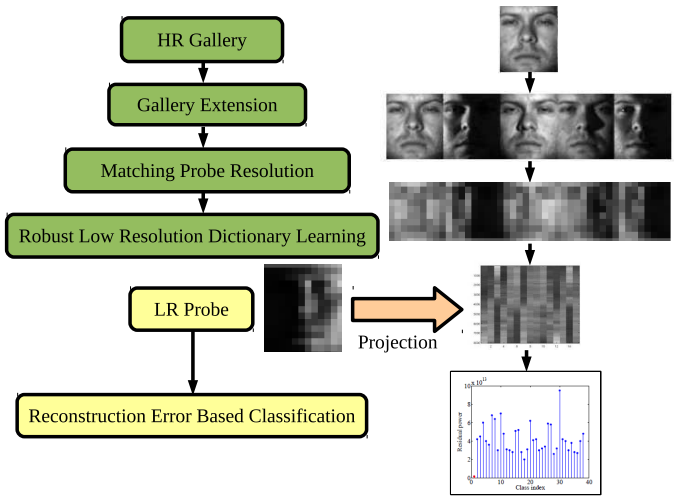}}
\caption{Overview of our algorithm.}
\label{fig:AlgDemo}
\end{figure}

Training stage of our method consists of three main steps.  In the first step of the training stage, given HR training samples from each class, we use an image relighting method to generate multiple images of the same subject with different lighting so that robustness to illumination changes can be realized.  In the second step, the resolution of the enlarged gallery images from each class is matched with that of the probe image.  Finally, in the third step, class and resolution specific dictionaries are trained for each class.  For the testing phase, a novel LR image is projected onto the span of the atoms in each learned dictionary.  The residual vectors are then used to classify the subject. A flowchart of the proposed algorithm is shown in figure~\ref{fig:AlgDemo}.

A preliminary version of this work appeared in \cite{ShekharIJCB2011}.  Extensions to \cite{ShekharIJCB2011} include kernalization of the dictionary learning algorithm as well as additional experiments using this kernalized algorithm.

\subsection{Paper organization}
The rest of the paper is organized as follows: In Section~\ref{sec:previous_work}, we review a few related works. In Section ~\ref{sec:ProposedApproach}, the proposed approach is described and in Section ~\ref{sec:Exp}, experimental results are demonstrated.  Finally, Section~\ref{sec:conclusion} concludes the paper with a brief summary and discussion.

\section{Previous Work} \label{sec:previous_work}
In this section, we review some of the recent FR methods that can deal with poor resolution.  These techniques can be broadly divided into the following categories.

\subsection{SR-based approaches}
SR is the method of estimating HR image $\mathbf{x}$ given downgraded image $\mathbf{y}$.  The LR image model is often given as $$\mathbf{y}=\mathbf{BHx} + \eta,$$ where $\mathbf{B}, \mathbf{H}$ and $\eta$ are the downsampling matrix, the blurring matrix and the noise, respectively.  Earlier works for solving the above problem were based on taking multiple LR inputs and combining them to produce the HR image. A classical work by Simon and Baker \cite{BakerSR} showed that the methods using multiple LR images using smooth priors would fail to produce good results as the resolution factor increases.  They also proposed a face hallucination method for super-resolving face images.  Subsequently, there have been works using single image for SR such as example-based SR \cite{BillFreeSR}, SR using neighborhood embedding \cite{HongSR} and sparse representation-based SR \cite{YiMaSR}.  While these methods can be used for super-resolving the face images and subsequent recognition, methods have also been proposed for specifically handling the problem for faces.

In particular, an eigen-face domain SR method for FR was proposed by Gunturk \emph{et al.} in \cite{EigSRBah}. This method proposes to solve the FR at LR using SR of multiple LR images using their PCA domain representation.  Given an LR face image, Jia and Gong \cite{TensorFaceSR} propose to directly compute a maximum likelihood identity parameter vector in the HR tensor space that can be used for SR and recognition.  Hennings-Yeomans \emph{et al.} \cite{S2R2Henning} presented a Tikhonov regularization method that can combine the different steps of SR and recognition in one step. Wilman \emph{et al.} \cite{WilmanTIP2012} proposed a relational learning approach for super-resolution and recognition of low resolution faces. 

\subsection{Metric learning-based approaches}
Though the LR are directly not suitable for face recognition purpose, it is also not necessary to super-resolve the image before recognition, as the problem of recognition is not the same as SR. Based on this motivation, some different approaches to this problem have been suggested. Coupled Metric Learning \cite{Xilin2010} attempts to solve this problem by mapping the LR image to a new subspace, where higher recognition can be achieved. A similar approach for improving the matching performance of the LR images using multidimensional scaling was recently proposed by Biswas \emph{et al.} in \cite{BiswasBTAS2010, BiswasPAMI2013, BiswasPAMI2012}. Further, Ren \emph{et al.} \cite{RenTIP2012} used coupled kernel methods for low resolution recognition. A coupled Fisher analysis method was proposed by Sienna \emph{et al} \cite{SienaECCV2012}. Lei \emph{et al} \cite{LeiTIFS2012}. also proposed a coupled discriminant analysis framework for heterogenous face recognition. 

\subsection{Other methods}
There have been works to solve the problem of unconstrained FR using videos.  In particular, Arandjelovic and Cipolla \cite{ShapeIllumMan} use a video database of LR face images with a variability in pose and illumination.  Their method combines a photometric model of image formation with a statistical model of generic face appearance variation to deal with illumination.  To handle pose variation, it learns local appearance manifold structure and a robust same-identity likelihood.

A change in resolution of the image changes the scale of the image.  Scale change has a multiplicative effect on the distances in image.  Hence, if the image is represented in log-polar domain, a scale change will lead to a translation in the said domain.  Based on this, a FR approach has been suggested by Hotta \emph{et al.} in \cite{LogPolar} to make the algorithm scale invariant.  This method proposes to extract shift-invariant features in the log-polar domain.

Additional methods for LR FR include correlation filter-based approach \cite{KerCor} and a support vector data description method \cite{SupportVectorLR}. 3D face modelling has also been used to address the LR face recognition problem \cite{MedioniNonCoop} \cite{SparseStereo}. Choi \emph{et al} \cite{ChoiTIC2009} make an interesting study on the use of color for degraded face recognition.

\section{Proposed Approach} \label{sec:ProposedApproach}
In this section, we present the details of our proposed low-resolution FR algorithm based on learning class specific dictionaries. 

\begin{figure*}[htp!]
\centering
\fbox{\includegraphics[width = 0.6\textwidth, height = 0.33\textwidth]{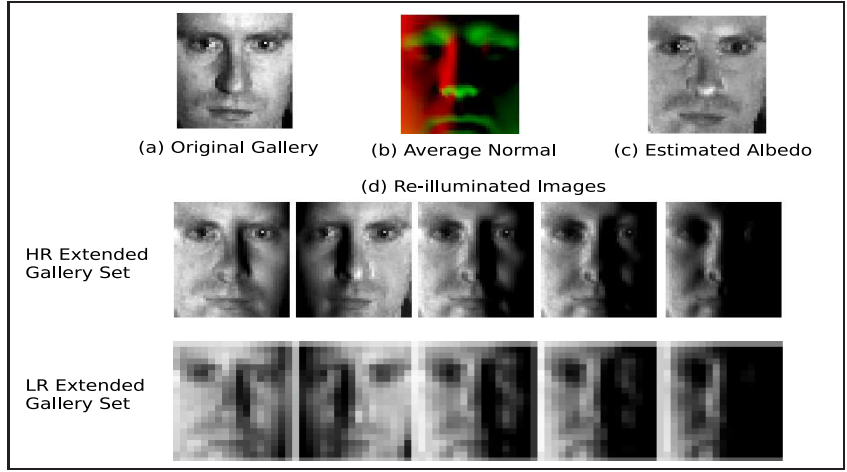}}
\caption{Examples of the (a) original image, (b) average normal used for calculation, (c) estimated albedo and (d) re-illuminated HR and LR gallery images.}
\label{fig:Relight}
\end{figure*}

\subsection{Image Relighting}\label{sec:relighting}
As discussed earlier, the resolution change is usually coupled with other parameters such as illumination variation.  In this section, we introduce an image relighting method that can deal with this illumination problem in LR face recognition. The idea is to capture various illumination conditions using the HR training samples, and subsequently use the expanded gallery for recognition at low resolutions.

Assuming the Lambertian reflectance model for facial surface, the HR intensity image $\mathbf{X}^H$ is given by the Lambert's cosine law as follows:
\begin{equation}
\mathbf{X}^H_{i,j}=\mathbf{\rho_{i,j}}\max(\mathbf{n}^{T}_{i,j}\mathbf{s},0),
\label{eq:lambert}
\end{equation}
where $\mathbf{X}^H_{i,j}$ is the pixel intensity at location $(i,j)$, $\mathbf{s}$ is the light source direction, $\rho(i,j)$ is the surface albedo at location $(i,j)$, $\mathbf{n} (i,j)$ is the surface normal of the corresponding surface point. Given the face image, $\mathbf{X}_H$,  image relighting involves estimating $\mathbf{\rho}$, $\mathbf{n}$ and $\mathbf{s}$, which is an extremely ill-posed problem. To overcome this, we use 3D facial normal data \cite{Blanz} to  first estimate an average surface normal, $\mathbf{\bar{n}}$. Further, the model is non-linear due to the $\max$ term in  (\ref{eq:lambert}). However, the shadow points do not reveal any information about albedo. Hence, we neglect the $\max$ term in further discussion. The albedo, $\mathbf{\rho}$ and source directions $\mathbf{s}$ can now be estimated as follows:

\begin{itemize}
\item  The source direction can be estimated using $\mathbf{\bar{n}}$ following a linear Least Squares approach \cite{HornBook}:
$$\mathbf{\hat{s}} = \left(\sum_{i,j} \mathbf{\bar{n}}_{i,j} \mathbf{\bar{n}}_{i,j}^T \right)^{-1} \sum_{i,j}
\mathbf{X}^H_{i,j} \mathbf{\bar{n}}_{i,j}$$
\item An inital estimate of albedo, $\mathbf{\rho}^{0}$ can be obtained as:
$$\mathbf{\rho}^{0}_{i,j} = \frac{\mathbf{X}^H_{i,j}}{\mathbf{\bar{n}}_{i,j}^T\mathbf{\hat{s}}}$$
\item The final albedo estimate is obtained using minimum mean square approach based on Wiener filtering framework \cite{SomaIllum}:
$$\mathbf{\hat{\rho}} = E(\mathbf{\rho}|\mathbf{\rho}^{0})$$
where, $E(\mathbf{\rho}|\mathbf{\rho}^{0})$ denotes the minimum mean square estimate (MMSE) of the albedo.
\end{itemize}  

Using the estimated albedo map, $\mathbf{\hat{\rho}}$ and average normal, $\mathbf{\bar{n}}$ we can generate new images under any illumination condition using the image formation model (\ref{eq:lambert}). It was shown in \cite{Lee9pts} that an image of an arbitrarily illuminated object can be approximated by a linear combination of the image of the same object in the same pose, illuminated by nine different light sources placed at preselected positions.  

Hence, the image formation equation can be rewritten as
\begin{equation}
\mathbf{X}=\sum_{k=1}^{9}a_{k}\mathbf{X}_{k},
\label{eq:9eq}
\end{equation}
where $$\mathbf{X}_{k}=\rho\max(\mathbf{n}^{T}\mathbf{s}_{i},0),$$ and $\{\mathbf{s}_{1},\cdots,\mathbf{s}_{9}\}$ are pre-specified illumination directions. Since, the objective is to generate HR gallery images which will be sufficient to account for any illumination in the probe image, we generate images under pre-specified illumination conditions and use them in the gallery.  Figure~\ref{fig:Relight} shows some relighted HR images along with the corresponding input and LR images. Furthermore, as the condition is true irrespective of the resolution of LR image, the same set of gallery images can be used for all resolutions.

\subsection{Low Resolution Dictionary Learning}
In LR face recognition, given labeled HR training images, the objective is to identify the class of a novel probe LR face image. Suppose that we are given $C$ distinct face classes and a set of $m_{i}$ HR training images per class, $i=\{1,\cdots,C\}$.  Here, $m_{i}$ corresponds to the total number of images in class $i$ including the relighted images. We identify an $l_H \times q_H$ grayscale image as an $N_H$-dimensional vector, $\mathbf{x}_H$, which can be obtained by stacking its columns, where $N_H =r_H \times q_H$.  Let $$\mathbf{X}^{H}_{i}=[\mathbf{x}^{H}_{i1},\cdots,\mathbf{x}^{H}_{im_{i}}]\in\R^{N_H \times m_{i}}$$ be an $N_H \times m_{i}$ matrix of training images corresponding to the $i^{th}$ class. For resolution and illumination robust recognition, the matrix $\mathbf{X}^{H}_{i}$ is pre-multiplied by downsampling $\mathbf{B}$ and blurring $\mathbf{H}$ matrices. Here, $\mathbf{H}$ has a fixed dimension of $N_H \times N_H$ and $\mathbf{B}$ will be of size $N_L \times N_H$, where $N_L = r_L \times q_L$, the LR probe being a grayscale image of $r_L \times q_L$. The resolution specific training matrix, $\mathbf{X}^{L}_{i}$ is thus created as
\begin{equation} \label{eq:LR_Dict_Mat}
\mathbf{X}^{L}_{i} = \mathbf{B} \mathbf{H} \mathbf{X}^{H}_{i} \triangleq (\mathbf{X}^{H}_{i})\downarrow.
\end{equation}

Given this matrix, we seek the dictionary that provides the best representation for each elements in this matrix.  One can obtain this by finding a dictionary $\mathbf{D}_{i}$ and a sparse matrix $\boldsymbol{\Gamma}_{i}$ that minimizes the following representation error
\begin{align}\label{mod_ksvd}
\nonumber (\mathbf{\hat{D}_{i}},\hat{\boldsymbol{\Gamma}}_{i})=\argmin{\mathbf{D}_{i},\boldsymbol{\Gamma}_{i}}\|\mathbf{X}^{L}_{i}&-\mathbf{D}_{i}\boldsymbol{\Gamma}_{i}\|_{F}^{2}
\textrm{\;\;subject to}\\&\;\|\boldsymbol{\gamma}_{k}\|_{0}\leq T_{0} \;\;\; \forall k,
\end{align}
where $\boldsymbol{\gamma}_{k}$ represent the columns of $\boldsymbol{\Gamma}_{i}$ and the $\ell_{0}$ sparsity measure $\|.\|_{0}$ counts the number of nonzero elements in the representation.  Here, $\|\mathbf{A}\|_{F}$ denotes the Frobenius norm defined as $\|\mathbf{A}\|_{F}=\sqrt{\sum_{i}\sum_{j}|\mathbf{A}(i,j)|^{2}}$.  Many approaches have been proposed in the literature for solving such optimization problem. In this paper, we adapt the K-SVD algorithm \cite{EladKSVD} for solving (\ref{mod_ksvd}) due to its simplicity and fast convergence.  The K-SVD algorithm alternates between sparse-coding and dictionary update steps.  In the sparse-coding step, $\mathbf{D}_{i}$ is fixed and the representation vectors $\boldsymbol{\gamma}_{k}$s are found for each example $\mathbf{x}^L_{i}$. Then, with fixed a $\boldsymbol{\Gamma}_{i},$ the dictionary is updated atom-by-atom in an efficient way.  See \cite{EladKSVD} for more details on the K-SVD dictionary learning algorithm.

\textbf{Classification:} 
Given an $r_L \times q_L$  LR probe, it is column-stacked to give the column vector  $\mathbf{y}$. It is projected onto the span of the atoms in each $\mathbf{D}_{i}$ of the $C$ class dictionary, using the orthogonal projector
$$\mathbf{P}_{i}=\mathbf{D}_{i}(\mathbf{D}_{i}^{T}\mathbf{D}_{i})^{-1}\mathbf{D}_{i}^{T}.$$
The approximation and residual vectors can then be calculated as
\begin{equation}\hat{\mathbf{y}}_{i}=\mathbf{P}_{i}\mathbf{y}
=\mathbf{D}_{i}\boldsymbol{\alpha}_{i}\label{eq:lin_approx}
\end{equation} 
and
\begin{align}\nonumber \mathbf{r}_{i}(\mathbf{y})& =\mathbf{y}-\mathbf{\hat{y}}_{i}\nonumber \\&=(\mathbf{I}-\mathbf{P}_{i})\mathbf{y},
\label{eq:lin_residual}
\end{align}

respectively, where $\mathbf{I}$ is the identity matrix and
\begin{equation}
\boldsymbol{\alpha}_{i}=(\mathbf{D}_{i}^{T}\mathbf{D}_{i})^{-1}\mathbf{D}_{i}^{T}\mathbf{y}
\label{eq:lin_ls}
\end{equation}
are the coefficients. Since the K-SVD algorithm finds the
dictionary, $\mathbf{D}_{i}$, that leads to the best
representation for each examples in $\mathbf{X}^{L}_{i}$, $\|\mathbf{r}_{i}(\mathbf{y})\|_{2}$ will be small if $\mathbf{y}$
were to belong to the $i^{th}$ class and large for the other
classes. Based on this, we can classify $\mathbf{y}$ by assigning
it to the class, $d\in\{1,\cdots,C\}$, that gives the lowest
reconstruction error, $\|\mathbf{r}^{i}(\mathbf{y})\|_{2}$:
\begin{align}
\nonumber d&=\textrm{identity}(\mathbf{y})\\&=\arg\min_{i}
\|\mathbf{r}_{i}(\mathbf{y})\|_{2}.
\label{eq:lin_id}
\end{align}

\textbf{Generic Dictionary Learning:}
The class-specific dictionary, $\mathbf{D}_{i}, i = 1, \cdots, C$ learnt above can be extended to use features other than intensity images. Specifically, the dictionary can be learnt using features like Eigenbasis,  $\mathbf{F}^{H}_{i}$ extracted from training matrix $\mathbf{X}^{H}_{i}$. However, as equation ~(\ref{eq:LR_Dict_Mat}) does not hold for $\mathbf{F}^{H}_{i}$, the resolution specific  feature matrix $\mathbf{F}^{L}_{i}$ is directly extracted using $\mathbf{X}^{L}_{i}$.
Our Synthesis-based LR FR (SLRFR) algorithm is summarized in Figure~\ref{Alg:SLRFR}.


\begin{figure}[htp!]
\centering
\newlength{\fiiil}
\setlength{\fiiil}{0.47\textwidth}
\addtolength{\fiiil}{-3\fboxsep} \addtolength{\fiiil}{-3\fboxrule}
\fbox{
\begin{minipage}{\fiiil}
Given a LR test sample $\mathbf{y}$ and $C$ training matrices
$\{ {\mathbf{X}}^H_{i} \}_{i=1}^C$ corresponding to HR gallery images. \\

{\bf{Procedure:}}
\begin{itemize}

\item For each training image, use the relighting approach described
in section~\ref{sec:relighting} to generate multiple images with
different illumination conditions and use them in the gallery.

\item Learn the best dictionaries $\mathbf{D}_{i}$, to represent the
resolution specific enlarged training matrices, $\mathbf{X}^{L}_{i}$,
using the K-SVD algorithm, where $\mathbf{X}^{L}_{i} = (\mathbf{X}^{H}_{i})\downarrow$, $i=1,\cdots,C$.

\item Compute the approximation vectors, $\mathbf{\hat{y}}^{i}$, and
the residual vectors, $\mathbf{r}^{i}(\mathbf{y})$, using (\ref{eq:lin_approx}) and (\ref{eq:lin_residual}), respectively for $i=1,\cdots,C$.

\item Identify $\mathbf{y}$ using (\ref{eq:lin_id}).
\end{itemize}

\end{minipage}}
\caption{The SLRFR algorithm.}\label{Alg:SLRFR}
\end{figure}

\subsection{Non-linear Dictionary Learning}
The class identities in the face dataset may not be linearly separable. Hence, we also extend the SLRFR framework to the kernel space.  This essentially requires the dictionary learning model to be non-liner \cite{Hien_ICASSP2012}.  

Let $\boldsymbol \phi^L:  \mathbb R^{N_{L}} \rightarrow G$ be a non-linear mapping from $N_{L}$ dimensional space into a dot product space $G$.  A non-linear dictionary can be trained in the feature space $G$ by solving the following optimization problem
 \begin{align}
\nonumber (\hat{\mathbf{A}}_{i}, \hat{\boldsymbol{\Gamma}}_{i})=
\argmin{\mathbf{A}_{i},\boldsymbol{\Gamma}_{i}}\| \mathbf \phi^L (\mathbf{X}^L_{i})&-\boldsymbol{\phi}^L(\mathbf{X}^L_{i})\mathbf{A}_{i}\boldsymbol{\Gamma}_{i}\|_{F}^{2}
\textrm{\;\;subject to} \\&\; \|\boldsymbol{\gamma}_{k}\|_{0}\leq T_{0} \;\;\; \forall k
\label{eq:ker_mod_ksvd}
\end{align}
where $$\boldsymbol{\phi}^L(\mathbf{X}^L_{i})=[\boldsymbol{\phi^L}(\mathbf{x}^L_{i1}), \cdots, \boldsymbol{\phi}^L(\mathbf{x}^L_{im_{i}})].$$ In (\ref{eq:ker_mod_ksvd}) we have used the following model for the dictionary in the feature space,
$$\tilde{\mathbf{D}}_{i}=\boldsymbol{\phi}^{L}(\mathbf{X}^{L}_{i})\mathbf{A}_{i},$$ 
Since it can be shown that the dictionary lies in the linear span of the samples $\boldsymbol{\phi}^L (\mathbf{X}^L_{i})$,
where $\mathbf{A}_{i} \in \mathbb{R}^{m_{i}\times K}$ is a matrix with $K$ atoms \cite{Hien_ICASSP2012}.
This model provides adaptivity via modification of the matrix $\mathbf{A}_{i}$.  Through some algebraic manipulations, the cost function in (\ref{eq:ker_mod_ksvd})
can be rewritten as,
 \begin{align}
\nonumber  \| \boldsymbol{\phi}^{L}(\mathbf{X}^{L}_{i}) -& \boldsymbol{\phi}^{L}(\mathbf{X}^{L}_{i}) \mathbf{A}_{i} \boldsymbol{\Gamma}_{i} \|_{F}^{2}\\&=\mathbf{tr} ( (\mathbf{I} - \mathbf{A}_{i}
\boldsymbol{\Gamma}_{i})^T \boldsymbol{\mathcal{K}}^{L}(\mathbf{X}^{L}_{i}, \mathbf{X}^{L}_{i})(\mathbf{I} - \mathbf{A}_{i} \boldsymbol{\Gamma}_{i})),
\end{align}
where $\boldsymbol{\mathcal{K}}^{L}$ is a kernel matrix whose elements are computed from $$\kappa(i,j) =\boldsymbol{\phi}^{L}(\mathbf{x}^{L}_i)^T \boldsymbol{\phi}^{L}(\mathbf{x}^{L}_j).$$ It is apparent that the objective function is feasible since it only involves a matrix of finite dimension $\boldsymbol{\mathcal{K}}^{L} \in \mathbb{R}^{m_{i}\times m_{i}}$, instead of dealing with a possibly infinite dimensional dictionary.

An important property of this formulation is that the computation
of $\boldsymbol{\mathcal{K}}^{L}$ only requires dot products. Therefore, we are
able to employ Mercer kernel functions to compute these dot
products without carrying out the mapping $\boldsymbol{\phi}^{L}$. Some commonly used kernels include polynomial kernels $$\kappa(\mathbf{x},\mathbf{y})=\langle \left(\mathbf{x},\mathbf{y}\rangle+c\right)^{d}$$ and Gaussian kernels $$\kappa(\mathbf{x},\mathbf{y})=\exp\left(-\frac{\|\mathbf{x}-\mathbf{y}\|^{2}}{c}\right),$$ where $c$ and $d$ are the parameters.

Similar to the optimization of (\ref{mod_ksvd}) using the linear K-SVD \cite{EladKSVD} algorithm, the optimization of (\ref{eq:ker_mod_ksvd}) involves sparse coding and dictionary update steps in the feature space which results in the kernel dictionary learning algorithm \cite{Hien_ICASSP2012}.  Details of the optimization can be found in \cite{Hien_ICASSP2012} and Appendix A.

\textbf{Classification:} 
Let $\{\mathbf{A}_{i}\}_{i=1}^{C}$ denote the learned dictionaries for $C$ classes.  Let $\mathbf{z} \in \R^{N_{L}}$ be a vectorized LR probe image $z$ of size $r_{L}\times q_{L}$.  We first find coefficient vectors $\boldsymbol{\gamma}_{i}\in\R^{K}$ with at most $T$ non-zero coefficients such that $\boldsymbol{\phi}^L(\mathbf{X}^{L}_{i})\mathbf{A}_{i}\boldsymbol{\gamma}_{i}$ approximates $\mathbf{z}$ by minimizing the following problem 
\begin{equation}
\underset{\boldsymbol{\gamma}_{i}}{\min} \;\; \|\boldsymbol{\phi}^{L}(\mathbf{z})-\boldsymbol{\phi}^{L}(\mathbf{X}^{L}_{i}) \mathbf{A}_{i} \boldsymbol{\gamma}_{i}\|^{2}_{2}
\;\;s.t\;\; \|\boldsymbol{\gamma}_{i}\|_{0}\le T,
\label{eq:subKKSVD}
\end{equation}
for all $i=1,\cdots,C$. The above problem can be solved by the Kernel Orthogonal Matching Pursuit (KOMP) algorithm \cite{Hien_ICASSP2012}.  The reconstruction error is then computed as
\begin{align}
\mathbf{r}_i & = \|\boldsymbol{\phi}^L(\mathbf{z})-\boldsymbol{\phi}^L(\mathbf{X}^L_{i}) \mathbf{A}_i \boldsymbol{\gamma}_{i} \|^2  \nonumber \\
& = \boldsymbol{\mathcal{K}}^{L}(\mathbf{z}, \mathbf{z}) - 2 \boldsymbol{\mathcal{K}}^{L}(\mathbf{z},\mathbf{X}_{i}) \mathbf{A}_i \boldsymbol{\gamma}_{i} +  \boldsymbol{\gamma}_{i}^T \mathbf{A}_i^T \boldsymbol{\mathcal{K}}^{L}(\mathbf{X}_{i},\mathbf{X}_{i}) \mathbf{A}_i \boldsymbol{\gamma}_{i},
\label{eq:ker_residual}
\end{align}
where $\boldsymbol{\mathcal{K}}^{L}(\mathbf{z},\mathbf{X}^{L}_{i})=[\kappa(\mathbf{z},\mathbf{x}^{L}_{i_{1}}), \kappa(\mathbf{z},\mathbf{x}^{L}_{i_{2}}), \cdots, \kappa(\mathbf{z},\mathbf{x}^{L}_{i_{m_{i}}})].$  Similar to the linear case, once the residuals are found, we can classify $\mathbf{z}$ by assigning
it to the class, $d\in\{1,\cdots,C\}$, that gives the lowest
reconstruction error, $\|\mathbf{r}^{i}(\mathbf{y})\|_{2}$:
\begin{align}
\nonumber d&=\textrm{identity}(\mathbf{y})\\&=\arg\min_{i}
\|\mathbf{r}_{i}(\mathbf{y})\|_{2}.
\label{eq:kernel_id}
\end{align}  Our kernel Synthesis-based LR FR (kerSLRFR) algorithm is summarized in Figure~\ref{Alg:KSLRFR}.

\begin{figure}[htp!]
\centering
\setlength{\fiiil}{0.47\textwidth}
\addtolength{\fiiil}{-3\fboxsep} \addtolength{\fiiil}{-3\fboxrule}
\fbox{
\begin{minipage}{\fiiil}
Given a LR test sample $\mathbf{y}$ and $C$ training matrices
$\{ {\mathbf{X}}^H_{i} \}_{i=1}^C$ corresponding to HR gallery images. \\

{\bf{Procedure:}}
\begin{itemize}

\item For each training image, use the relighting approach described in section~\ref{sec:relighting} to generate multiple images with
different illumination conditions and use them in the gallery.

\item Learn non-linear dictionaries $\mathbf{A}_{i}$, to represent the
resolution specific enlarged training matrices, ${\mathbf{X}}^L_{i}$,
using the kernel dictionary learning algorithm \ref{eq:ker_mod_ksvd}, where ${\mathbf{X}}^L_{i} = (\mathbf{X}^H_{i})\downarrow$, $i=1,\cdots,C$.

\item Compute the sparse codes, $\mathbf{\gamma}_i$ and the residual vectors, $\mathbf{r}^{i}$, using (\ref{eq:subKKSVD}) and (\ref{eq:ker_residual}), respectively for $i=1,\cdots,C$.

\item Identify $\mathbf{y}$ using (\ref{eq:kernel_id}).
\end{itemize}

\end{minipage}}
\caption{The kerSLRFR algorithm.}\label{Alg:KSLRFR}
\end{figure}

\subsection{Joint Non-linear Dictionary Learning}
In the previous sections, we described methods to learn resolution-specific dictionaries for linear and non-linear cases. However, even though dictionaries can capture class-specific variations, the recognition performance would go down at low resolutions. Hence, information available in the HR training images must be exploited to make the method robust. To this, we propose a framework of learning joint dictionaries for HR and corresponding LR images. We achieve this through sharing sparse codes between HR and LR dictionaries. This regularizes the learned LR dictionary to output similar sparse codes as HR dictionary, thus, making it robust. The proposed formulation is described as follows.  

Let $\boldsymbol \phi^H:  \mathbb R^{N_{H}} \rightarrow G$ be a non-linear mapping from $N_{H}$ dimensional space into a dot product space $G$. We seek to learn dictionaries $\mathbf{A}^{H}$ and $\mathbf{A}^{L}$ by solving the optimization problem:
\begin{align}
\nonumber (\hat{\mathbf{A}}^H_{i}, \hat{\mathbf{A}}^L_{i}, \hat{\boldsymbol{\Gamma}}_{i}) &=
\argmin{\mathbf{A}^H_{i}, \mathbf{A}^L_{i}, \boldsymbol{\Gamma}_{i}} \| \mathbf \phi^H (\mathbf{X}^H_{i})- \boldsymbol{\phi}^H(\mathbf{X}^H_{i})\mathbf{A}^{H}_{i}\boldsymbol{\Gamma}_{i}\|_{F}^{2} \nonumber \\ 
& + \lambda \| \mathbf \phi^L (\mathbf{X}^L_{i})-\boldsymbol{\phi}^L(\mathbf{X}^L_{i})\mathbf{A}^{L}_{i}\boldsymbol{\Gamma}_{i}\|_{F}^{2} \nonumber \\& \textrm{ subject to }   \|\boldsymbol{\gamma}_{k}\|_{0}\leq T_{0} \text{ } \forall k
\label{eq:joint_ker_ksvd}
\end{align}  
where, $\lambda > 0$ is a hyperparameter. This can be re-formulated as: 
\begin{align}
\nonumber (\hat{\mathbf{\tilde{A}}}_{i}, \hat{\boldsymbol{\Gamma}}_{i}) &=
\argmin{\mathbf{\tilde{A}}, \boldsymbol{\Gamma}_{i}} \| \boldsymbol{\Phi}_1(\mathbf{X}^H_i, \mathbf{X}^L_i)- \boldsymbol{\Phi}_2(\mathbf{X}^H_{i}, \mathbf{X}^L_{i})\mathbf{\tilde{A}}_{i}\boldsymbol{\Gamma}_{i}\|_{F}^{2} \nonumber \\ 
& \textrm{ subject to }   \|\boldsymbol{\gamma}_{k}\|_{0}\leq T_{0} \text{ } \forall k
\label{eq:joint_ker_ksvd_reform}
\end{align}  
where,
$$\boldsymbol{\Phi}_1(\mathbf{X}^H_i, \mathbf{X}^L_i) = \left[ \begin{array}{c} \mathbf{\phi}(\mathbf{X}_H) \\ \sqrt{\lambda}\mathbf{\phi}(\mathbf{X}_L) \end{array} \right], \mathbf{\tilde{A}} = \left[ \begin{array}{c} \mathbf{A}^{H}_{i} \\ \mathbf{A}^{L}_{i} \end{array} \right]$$

$$\boldsymbol{\Phi}_1(\mathbf{X}^H_i, \mathbf{X}^L_i) = \left[ \begin{array}{cc} \mathbf{\phi}(\mathbf{X}_H) & \mathbf{0} \\ \mathbf{0} & \sqrt{\lambda}\mathbf{\phi}(\mathbf{X}_L) \end{array} \right]$$ 

The optimization problem (\ref{eq:joint_ker_ksvd}) can be solved in a similar way as (\ref{eq:ker_mod_ksvd}) using a modified version of kernel K-SVD algorithm \cite{Hien_ICASSP2012}. Details of the method are presented in Appendix A. 

\textbf{Classification: } Let $\{\mathbf{A}^{L}_{i}\}_{i=1}^{C}$ denote the learned dictionaries for $C$ classes. Then a low resolution probe $\mathbf{z} \in \R^{N_{L}}$ can be classified using the KOMP algorithm \cite{Hien_ICASSP2012}, as described in (\ref{eq:subKKSVD}), (\ref{eq:ker_residual}) and (\ref{eq:kernel_id}), by substituting $\{\mathbf{A}^{L}_{i}\}_{i=1}^{C}$ for dictionary term. The proposed algorithm joint kernel SLRFR (jointKerSLRFR) is summarized in Figure \ref{Alg:jointkerSLRFR}. 

\begin{figure}[htp!]
\centering
\setlength{\fiiil}{0.47\textwidth}
\addtolength{\fiiil}{-3\fboxsep} \addtolength{\fiiil}{-3\fboxrule}
\fbox{
\begin{minipage}{\fiiil}
Given a LR test sample $\mathbf{y}$ and $C$ training matrices
$\{ {\mathbf{X}}^H_{i} \}_{i=1}^C$ corresponding to HR gallery images. \\

{\bf{Procedure:}}
\begin{itemize}

\item For each training image, use the relighting approach described
in section~\ref{sec:relighting} to generate multiple images with
different illumination conditions and use them in the gallery.

\item Learn the dictionaries $\mathbf{A}^H_{i}$ and $\mathbf{A}^L_{i}$ to jointly represent the
HR and LR training matrices, $\mathbf{X}^{H}_{i}$ and $\mathbf{X}^{L}_{i}$, where ${\mathbf{X}}^L_{i} = (\mathbf{X}^H_{i})\downarrow$, $i=1,\cdots,C$, respectively
using the joint kernel dictionary algorithm.

\item Compute the the sparse codes, $\mathbf{\gamma}_i$ and the residual vectors, $\mathbf{r}^{i}$, using (\ref{eq:subKKSVD}) and (\ref{eq:ker_residual}) respectively for $i=1,\cdots,C$.

\item Identify $\mathbf{y}$ using (\ref{eq:kernel_id}).
\end{itemize}

\end{minipage}}
\caption{The jointKerSLRFR algorithm.}\label{Alg:jointkerSLRFR}
\end{figure}

\section{Experiments} \label{sec:Exp}
To demonstrate the effectiveness of our method, in this section, we present experimental results on various face recognition datasets. We deomonstrate the effectiveness of proposed recognition framework, as well as compared with metric learning \cite{BiswasBTAS2010, Xilin2010} and super-resolution \cite{WilmanTIP2012, S2R2Henning} based methods. For all the experiments, we learnt the dictionary elements using PCA features. 

\subsection{FRGC Dataset}
We also evaluated on Experiment 1 of the FRGC dataset \cite{FRGC-data}. It consists of $152$ gallery images, each subject having one gallery and $608$ probe images under controlled setting. A separate training set of $183$ images is also available which was used to learn the PCA basis.

\textbf{Implementation} The resolution of the HR image was fixed at $48 \times 40$ and probe images at resolutions of  $12 \times 10$, $10 \times 8$ and $7 \times 6$ were created by smoothening and downsampling the HR probe images. From each gallery image, 5 different illumination images were produced, which were flipped to give 10 images per subject. The experiments were done at resolutions of $10 \times 8$ and $7 \times 6$, thus validating the method across resolutions. We also tested the CLPM algorithm \cite{Xilin2010} and PCA performances on the expanded gallery to get a fair comparison. We also report the recognition rate for PCA using the original gallery image to demonstrate the utility of gallery extension at low resolutions. Results from other algorithms are also tabulated. We chose RBF kernel for tesing kerSLRFR and jointKerSLRFR and set $\lambda = 1$ for jointKerSLRFR. The kernel parameter, $c$ was obtained through cross-validation for both HR and LR data.

\begin{figure}[htp!]
\centering
\includegraphics[width = 0.5\textwidth]{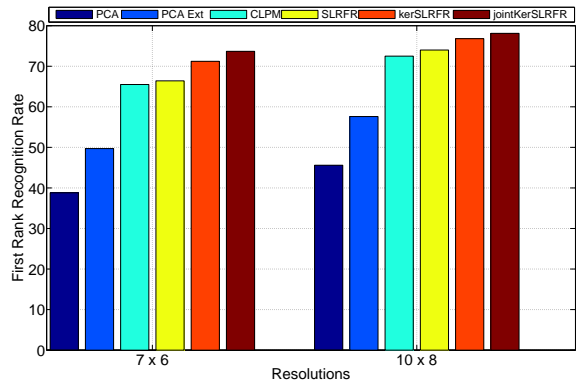}
\caption{Recognition Rates for FRGC data with probes at low resolutions}
\label{fig:FRGC_bar}
\end{figure}

\textbf{Observations} Figure \ref{fig:FRGC_bar} and Table \ref{tab:FRGC_prev} show that the proposed methods clearly outperforms previous algorithms. The proposed algorithm, SLRFR improves the CLPM algorithm for all the resolutions, while kerSLRFR further boosts the performance. The jointKerSLRFR shows the best performance for all the methods. The joint sparse coding framework, clearly helps in improving performance at low resolutions. Further, PCA using the extended gallery set also improves the performance over using a single gallery image. This shows that our method of gallery extension can be coupled with the existing face recognition algorithms to improve performance at low resolutions.

\begin{table*}[htp]
\centering
\begin{tabular}{c|c|c|c|c|c|c}
\hline
Resolution & MDS \cite{BiswasBTAS2010} & S2R2 \cite{S2R2Henning} & VLR \cite{WilmanTIP2012} & SLRFR & kerSLRFR & jointKerSLRFR \\
\hline
$6 \times 6$ &  - & $55.0 \%$ & - &   $62.9 \%$ & $64.7 \%$ & $\mathbf{65.2} \%$\\
$7 \times 6$ & - & - & $55.5 \%$ &  $63.8 \%$ & $71.2 \%$ & $\mathbf{73.6} \%$\\
$9 \times 7$ & $58.0 \%$ & - &  - &  $72.2 \%$ & $76.4 \%$ & $\mathbf{78.1} \%$\\
\hline
\end{tabular}
\caption{Comparisons for rank one recognition rate of FRGC dataset}
\label{tab:FRGC_prev}
\end{table*}

\begin{figure}[htp!]
\centering
\includegraphics[width = 0.5\textwidth]{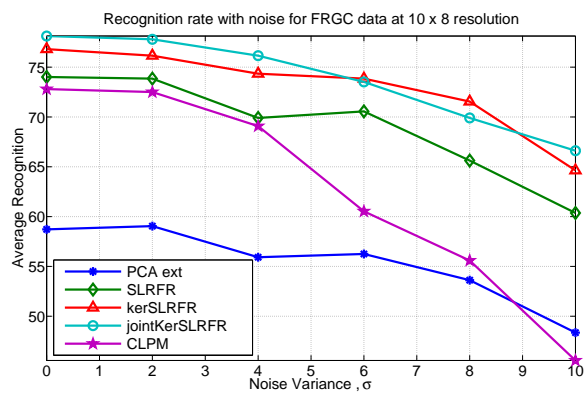}
\caption{Recognition Rates for FRGC data across increasing noise levels at $10 \times 8$ LR probe resolutions}
\label{fig:FRGC_noise}
\end{figure}

\textbf{Sensitivity to noise:} Low resolution images are often corrupted with noise. Thus, senstivity of noise is important in assessing performance of different algorithms. Figure \ref{fig:FRGC_noise} shows the recognition rate for different algorithms with increasing noise level. It can be seen that CLPM shows a sharp decline with increasing noise, but the proposed approaches SLRFR, kerSLRFR and jointKerSLRFR are stable with noise. This is because the CLPM algorithm learns a model tailored to noise-free low resolution images, whereas the generative approach in the proposed methods leads to stable performance with increasing noise. 

\subsection{CMU-PIE dataset}
The PIE dataset \cite{PIE-data} consists of $68$ subjects in frontal pose and under different illumination conditions. Each subject has $21$ face images under different illumination conditions.\\

\textbf{Implementation} We chose first $34$ subjects with $6$ randomly chosen illuminations as the training set to learn PCA basis. For the remaining $34$ subjects and the $15$ illumination conditions, the experiment was done by choosing one gallery image per subject and taking the remaining as the probe image. The procedure was repeated for all the images and the final recognition rate was obtained by averaging over all the images. The size of the HR images was fixed to $48 \times 40$.  The LR images were obtained by smoothening followed by downsampling the HR images. For each galley image, $10$ images under different illuminations produced using gallery extension method and the corresponding flipped images were added to the gallery set. The RBF kernel was chosen for kerSLRFR and jointKerSLRFR and the kernel parameter, $c$ was set through cross-validation.


\begin{figure}[htp]
\centering
\includegraphics[width = 0.5\textwidth]{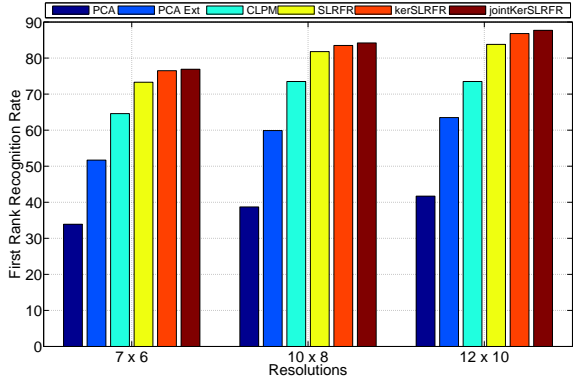}
\caption{Recognition Rates for PIE data with probes at low resolutions}
\label{fig:PIE_bar}
\end{figure}

\begin{figure}[htp!]
\centering
\includegraphics[width = 0.5\textwidth]{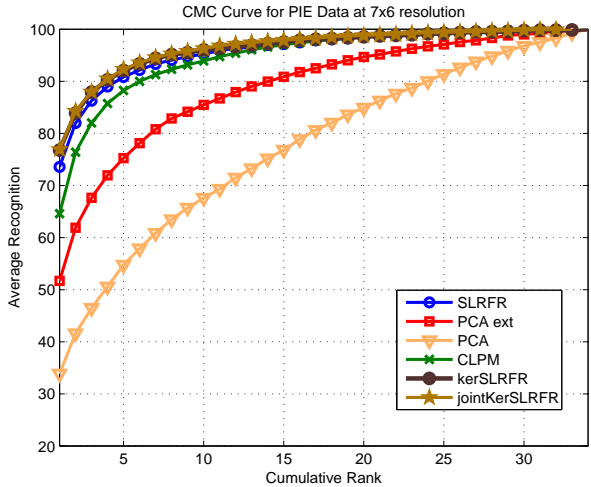}
\caption{CMC (Cumulative Match Characteristic) Curves for PIE data with probes at $7 \times 6$ resolution}
\label{fig:PIE_CMC}
\end{figure}

\begin{table*}[htp]
\centering
\begin{tabular}{c|c|c|c|c|c}
\hline
 Resolution & MDS \cite{BiswasBTAS2010} & VLR* \cite{WilmanTIP2012} & SLRFR & kerSLRFR & jointKerSLRFR  \\
\hline
 $7 \times 6$ & $55.0 \%$ & $74 \%$ & $73.3 \%$ & $76.5 \%$ & $\mathbf{76.9} \%$  \\
 $12 \times 10$ & $73.0 \%$ & - &  $83.8 \%$ & $86.8 \%$ & $\mathbf{87.4} \%$  \\
 $19 \times 16$ & $78.0 \%$ & - & $87.1 \%$ & $89.7 \%$ & $\mathbf{90.0} \%$ \\
\hline
\end{tabular}
\caption{Comparisons for rank one recognition of PIE dataset rate. Note that VLR* \cite{WilmanTIP2012} uses multiple gallery images while training.}
\label{tab:PIE_prev}
\end{table*}

\textbf{Observations}
Figure \ref{fig:PIE_bar}, \ref{fig:PIE_CMC} and Table \ref{tab:PIE_prev} show that the proposed method clearly outperforms previous algorithms. The proposed algorithms shows over $30 \%$ improvement over PCA performance with the original gallery set at rank one recognition rate and $8 \%$ better than the CLPM method at the lowest probe resolution. PCA using the extended gallery set also improves the performance over using a single gallery image. This shows that our method of gallery extension can be coupled with the existing face recognition algorithms to improve performance at low resolutions.

\subsection{AR Face dataset}
We also tested the proposed algorithms on the AR Face dataset \cite{ARFaceData}. TheAR face dataset consists of faces with varying illumination and expression conditions, captured
in two sessions. We evaluated our algorithms on a set of 100 users. Images from the first session, seven for each subject,were used as training and gallery and the images from the second
session, again seven per subject, were used for testing. \\

\textbf{Implementation} To test our method and compare with the existing metric learning based methods \cite{Xilin2010} \cite{BiswasBTAS2010}, we chose first $30$ subjects from the first session as the training set. For the remaining $70$ subjects, the experiment was done by choosing one gallery image per subject from the first session and taking the corresponding images from session 2 as probes. The procedure was repeated for all the $7$ images in the session 1 and the final recognition rate was obtained by averaging over all the runs. The size of the HR images was fixed to $55 \times 40$.  The LR images were obtained by smoothening followed by downsampling the HR images to $14 \times 10$. We also tested the CLPM algorithm \cite{Xilin2010} and PCA performances on the expanded gallery to get a fair comparison. Results from other algorithms are also tabulated.

\begin{figure}[htp!]
\centering
\includegraphics[width = 0.5\textwidth]{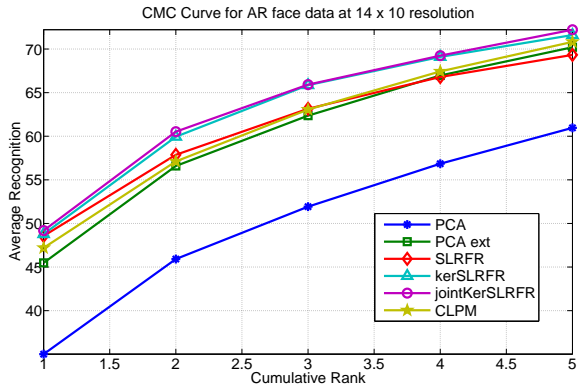}
\caption{CMC Curves for AR face data with probes at $14 \times 10$ resolution}
\label{fig:ar_face_CMC}
\end{figure}

\textbf{Observations} Figure \ref{fig:ar_face_CMC} shows the CMC curve for the first $5$ ranks. Clearly, the proposed approaches outperform other methods. SLRFR gives better rank one performance than CLPM algorithm, while kerSLRFR and jointKerSLRFR further increases the recognition over all the ranks.  

\subsection{Outdoor Face Dataset}
\begin{figure}[htp!]
\centering

\fbox{\includegraphics[width = 0.35\textwidth]{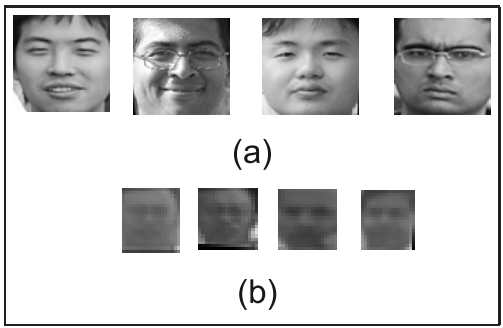}}
\caption{Example images from the outdoor face dataset (a) HR gallery images (b) LR probe images}
\label{fig:MURI_ex}
\end{figure}

We also tested our method on a challenging outdoor face dataset. The database consists of face images of $18$ individuals at different distances from camera. We chose a subset of $90$ low resolution images, which were also corrupted with blur, illumination and pose variations. $5$ high resolution, frontal and well-illuminated images were taken as the gallery set for each subject. The images were aligned using $5$ manually selected facial points. The gallery resolution was fixed at $120 \times 120$ and the probe resolution at $20 \times 20$. Figure \ref{fig:MURI_ex} shows some of the gallery images and the low quality probe images. The recognition rates for the dataset are shown in Table \ref{tab:MURI_rec}. We compare our method with the Regularized Discriminant Analysis (RDA) \cite{RegLDA} and CLPM \cite{Xilin2010}.  For the reg LDA comparison, we first used the PCA as a dimensionality reduction method to project the raw data onto an intermediate space, then we used the RDA to project the PCA coefficients onto a final feature space.

\begin{table}[htp]
\centering
\begin{tabular}{c|c}
\hline
Method & Recognition Rate\\
\hline
reg LDA & $60 \%$ \\
CLPM \cite{Xilin2010}	& $16.7 \%$ \\
SLRFR	 & $67.8 \%$ \\
kerSLRFR	 & $\mathbf{71.1} \%$ \\
jointKerSLRFR	 & $\mathbf{71.1} \%$ \\
\hline
\end{tabular}
\caption{Performance for the Outdoor Face Dataset}
\label{tab:MURI_rec}
\end{table}

\textbf{Observations} It can be seen from the table that SLRFR outperforms other algorithms on this difficult outdoor face dataset. The kerSLRFR algorithm further improves the performance, however, the jointKerSLRFR doesn't improve it further. This may be because this is a challenging dataset containing variations other than LR, like pose, blur, etc. The CLPM algorithm performs rather poorly on this dataset, as it is unable to learn the challenging variations in the dataset. 

\section{Computational Efficiency}
All the experiments were conducted using 2.13GHz Intel Xeon processor on Matlab programming interface. The gallery extension step using relighting took an average of $2s$  per gallery image of size $48 \times 40$. The K-SVD Dictionary took on an average $0.07s$ to train each class, while classification of a probe image was done in an average of $0.1s$ at the resolution of $7 \times 6$. Thus, the proposed algorithm is computationally efficient . Further, as the extended gallery can be used for all resolutions, it can be computed once and stored for a database.

\section{Discussion and Conclusion}\label{sec:conclusion}
We have proposed an algorithm which can provide good accuracy for low resolution images, even when a single HR gallery image is provided per person. While the method avoids the complexity of previously proposed algorithms, it is also shown to provide state-of-the-art results when the LR probe differ in illumination from the given gallery image. The idea of exploiting information in HR gallery image is novel and can be used to extend the limits of remote face recognition. Future extensions to this work will be to extend the proposed method to account for other variations such as pose, expression, etc. The present classification using reconstruction error can be studied further to explore a mix of discriminative and reconstructive techniques to further improve the recognition.

\section*{Acknowledgment}
This work was supported by a Multidisciplinary University Research Initiative grant from the Office of Naval Research under the grant N00014-08-1-0638.

\bibliographystyle{IEEEtran}
\bibliography{low_res_cit}

\section*{Appendix A}
Here, we will describe the kernel dictionary learning algorithm \cite{Hien_ICASSP2012} and the framework for the proposed joint kernel dictionary learning algorithm (jointKerKSVD). 

\subsection{Kernel Dictionary Learning}
The optimization problem (\ref{eq:ker_mod_ksvd}) can be solved in two stages.
\\ \\
\textbf{Sparse Coding:} Here, $\mathbf{A}_i$ is kept fixed while searching for the optimal sparse code, $\mathbf{\Gamma_i}$. The cost term in (\ref{eq:ker_mod_ksvd}) can be written as:
$$\| \mathbf \phi^L (\mathbf{X}^L_{i})-\boldsymbol{\phi}^L(\mathbf{X}^L_{i})\mathbf{A}_{i}\boldsymbol{\Gamma}_{i}\|_{F}^{2} = $$
$$ \sum_{j=1}^{m_i} \| \mathbf \phi^L (\mathbf{x}^L_{i,j})-\boldsymbol{\phi}^L(\mathbf{X}^L_{i})\mathbf{A}_{i}\boldsymbol{\Gamma}_{i, j}\|_{F}^{2}$$ 
Hence, the optimization problem can be broken up into $m_i$ different sub-problems:
$$\argmin{\mathbf{\Gamma}_{i,j}} \| \boldsymbol{\phi}^L (\mathbf{x}^L_{i,j})-\boldsymbol{\phi}^L(\mathbf{X}^L_{i})\mathbf{A}_{i}\boldsymbol{\Gamma}_{i, j}\|_{F}^{2}$$
We can solve this using kernel orthogonal matching pursuit (KOMP). Let $I_k$ denote the set of selected atoms at iteration $k$, $\mathbf{\hat{x}}_{k}$ denote the reconstruction of the signal, $\mathbf{x}^L_{i,j}$ using the selected atoms, $\mathbf{r}_{k}$ being the corresponding residue and $\mathbf{\gamma}_{k}$ the sparse code at $k^{th}$ iteration.
\begin{enumerate}
\item Start with $I_0 = \emptyset$, $\mathbf{\hat{x}}_{k} = 0$, $\mathbf{\gamma}_{k} = 0$.

\item Calculate the residue as:
 $$\mathbf{\phi}^L(\mathbf{x}^L_{i,j}) = \mathbf{\phi}^L(\mathbf{X}^L_{i})\hat{\mathbf{x}}_k + \mathbf{r}_k$$. 
 \item Project the residue on atoms not selected and add the atom with maximum projection value to $I_k$:
 \begin{align}
 \tau_t & = (\mathbf{\phi}^L(\mathbf{x}^L_{i,j}) - \mathbf{\phi}^L(\mathbf{X}^L_{i})\hat{\mathbf{x}}_k)^T(\mathbf{X}^L_{i}\mathbf{a}_t) \nonumber \\
& = (\boldsymbol{\mathcal{K}}^L(\mathbf{x}^L_{i,j}, \mathbf{X}^L_{i}) - \hat{\mathbf{x}}_k^T\boldsymbol{\mathcal{K}}^L(\mathbf{X}^L_{i},\mathbf{X}^L_{i}))\mathbf{a}_t, \; t \notin I_k
 \end{align}
Update the set $I_k$ as:
$$I_{k+1} = I_k \cup \argmax{t \notin I_k} |\tau_t|$$    
\item Update the sparse code, $\mathbf{\gamma}_{k+1}$ and reconstruction, $\mathbf{\hat{x}}_{k+1}$ as:
\begin{align}
\mathbf{\gamma}_{k+1} &= ((\mathbf{\phi}^L(\mathbf{X}^L_i)\mathbf{A}_{I_{k+1}})^T
(\mathbf{\phi}^L(\mathbf{X}^L_i)\mathbf{A}_{I_{k+1}}))^{-1} \nonumber \\ &(\mathbf{\phi}^L(\mathbf{X}^L_i)\mathbf{A}_{I_{k+1}})^T\mathbf{\phi}^L(\mathbf{x}^L_{i,j}) \nonumber \\
& = (\mathbf{A}^T_{I_{k+1}} \boldsymbol{\mathcal{K}}^L(\mathbf{X}^L_i,\mathbf{X}^L_i)\mathbf{A}_{I_{k+1}})^{-1} \nonumber \\
& (\boldsymbol{\mathcal{K}}^L(\mathbf{x}^L_{i,j},\mathbf{X}^L_i)\mathbf{A}_{I_{k+1}})^T \\
\hat{\mathbf{x}}_{k+1} & = \mathbf{A}_{I_{k+1}} \mathbf{\gamma}_{k+1}
\end{align}
\item $k \leftarrow k + 1$; Repeat steps 2-4 $T_0$ times.
\end{enumerate} 

\textbf{Dictionary update}
Once the sparse codes are calculated, the dictionary $\mathbf{A}_i$ can be updated as:
$$\mathbf{A}_i = \mathbf{\Gamma}_i^T(\mathbf{\Gamma}_i\mathbf{\Gamma}_i^T)^{-1}.$$ The dictionary atoms are now normalized to unit norm in feature space:
$$\mathbf{A}_{i,j} = \frac{\mathbf{A}_{i,j}}{\sqrt{\mathbf{A}_{i,j}^T \boldsymbol{\mathcal{K}}^L(\mathbf{X}^L_{i},\mathbf{X}^L_{i})\mathbf{A}_{i,j}}}, \; j = 1, \cdots, K$$   

\subsection{Joint kernel dictionary learning}
The optimization problem (\ref{eq:joint_ker_ksvd}) can be solved in a similar way as the kernel dictionary learning problem in two alterative steps:
\\ \\
\textbf{Sparse Coding}
Here, we keep $\mathbf{A}^H_i$ and $\mathbf{A}^L_i$ fixed and learn the joint sparse code $\mathbf{\Gamma}_i$. The optimization problem (\ref{eq:joint_ker_ksvd_reform}) can be written as:
$$\| \boldsymbol{\Phi}_1(\mathbf{X}^H_i, \mathbf{X}^L_i)- \boldsymbol{\Phi}_2(\mathbf{X}^H_{i}, \mathbf{X}^L_{i})\mathbf{\tilde{A}}_{i}\boldsymbol{\Gamma}_{i}\|_{F}^{2} = $$
$$ \sum_{j=1}^{m_i} \| \boldsymbol{\Phi}_1(\mathbf{X}^H_{i,j}, \mathbf{X}^L_{i,j})- \boldsymbol{\Phi}_2(\mathbf{X}^H_{i}, \mathbf{X}^L_{i})\mathbf{\tilde{A}}_{i}\boldsymbol{\Gamma}_{i,j}\|_{F}^{2} $$
Thus, the optimization can be broken up into $m_i$ sub-problems:
$$\argmin{\mathbf{\Gamma}_{i,j}}  \| \boldsymbol{\Phi}_1(\mathbf{X}^H_{i,j}, \mathbf{X}^L_{i,j})- \boldsymbol{\Phi}_2(\mathbf{X}^H_{i}, \mathbf{X}^L_{i})\mathbf{\tilde{A}}_{i}\boldsymbol{\Gamma}_{i,j}\|_{F}^{2} $$
This is similar to the original kernel dictionary learning formulation, with the signal $\mathbf{x}^L_{i,j})$ replaced by $\boldsymbol{\Phi}_1(\mathbf{X}^H_{i,j}, \mathbf{X}^L_{i,j})$. Thus, the above problem can be solved using similar procedure as KOMP. Let $I_k$ denote the set of selected atoms at iteration $k$, $\mathbf{\hat{x}}^{H,L}_{k}$ denote the reconstruction of the signal, $\boldsymbol{\Phi}_1(\mathbf{X}^H_{i,j}, \mathbf{X}^L_{i,j})$ using the selected atoms, $\mathbf{r}_{k}$ being the corresponding residue and $\mathbf{\gamma}_{k}$ the sparse code at $k^{th}$ iteration.

\begin{enumerate}
\item Start with $I_0 = \emptyset$, $\mathbf{\hat{x}}^{H,L}_{k} = 0$, $\mathbf{\gamma}_{k} = 0$.

\item Calculate the residue as:
 $$\boldsymbol{\Phi}_1(\mathbf{X}^H_{i,j}, \mathbf{X}^L_{i,j}) = \boldsymbol{\Phi}_2(\mathbf{X}^H_{i}, \mathbf{X}^L_{i})\hat{\mathbf{x}}^{H,L}_k + \mathbf{r}_k$$. 
 \item Project the residue on atoms not selected and add the atom with maximum projection value to $I_k$:
 \begin{align}
 \tau_t & = (\boldsymbol{\Phi}_1(\mathbf{X}^H_{i,j}, \mathbf{X}^L_{i,j}) - \boldsymbol{\Phi}_2(\mathbf{X}^H_{i}, \mathbf{X}^L_{i})\hat{\mathbf{x}}^{H,L}_k)^T \nonumber \\
 & ( \boldsymbol{\Phi}_2(\mathbf{X}^H_{i}, \mathbf{X}^L_{i})\mathbf{a}_t) \nonumber \\
& = (\boldsymbol{\mathcal{K}}^1 - (\hat{\mathbf{x}}_k^{H,L})^T\boldsymbol{\mathcal{K}}^2)\mathbf{\tilde{a}}_t, \; t \notin I_k
 \end{align}
where,
$$\boldsymbol{\mathcal{K}}^1 = \boldsymbol{\Phi}_1(\mathbf{X}^H_{i,j}, \mathbf{X}^L_{i,j})^T \boldsymbol{\Phi}_1(\mathbf{X}^H_{i,j}, \mathbf{X}^L_{i,j})$$
$$= \left[ \begin{array}{c}
\boldsymbol{\mathcal{K}}_H \\ \lambda \boldsymbol{\mathcal{K}}_L 
\end{array} \right]$$

$$\boldsymbol{\mathcal{K}}^2 = \boldsymbol{\Phi}_2(\mathbf{X}^H_{i,j}, \mathbf{X}^L_{i,j})^T \boldsymbol{\Phi}_2(\mathbf{X}^H_{i,j}, \mathbf{X}^L_{i,j})$$
$$= \left[ \begin{array}{cc}
\boldsymbol{\mathcal{K}}_H & \mathbf{0} \\
\mathbf{0} & \lambda \boldsymbol{\mathcal{K}}_L
\end{array}  \right]$$

Update the set $I_k$ as:
$$I_{k+1} = I_k \cup \argmax{t \notin I_k} |\tau_t|$$    
\item Update the sparse code, $\mathbf{\gamma}_{k+1}$ and reconstruction, $\mathbf{\hat{x}}_{k+1}$ as:
\begin{align}
\mathbf{\gamma}_{k+1} &= ((\boldsymbol{\Phi}_2(\mathbf{X}^H_{i,j}, \mathbf{X}^L_{i,j})\mathbf{\tilde{A}}_{I_{k+1}})^T
(\boldsymbol{\Phi}_2(\mathbf{X}^H_{i,j}, \mathbf{X}^L_{i,j})\mathbf{\tilde{A}}_{I_{k+1}}))^{-1} \nonumber \\ &(\boldsymbol{\Phi}_2(\mathbf{X}^H_{i,j}, \mathbf{X}^L_{i,j})\mathbf{\tilde{A}}_{I_{k+1}})^T\boldsymbol{\Phi}_1(\mathbf{X}^H_{i,j}, \mathbf{X}^L_{i,j}) \nonumber \\
& = (\mathbf{\tilde{A}}^T_{I_{k+1}} \boldsymbol{\mathcal{K}}^2\mathbf{\tilde{A}}_{I_{k+1}})^{-1}(\boldsymbol{\mathcal{K}}^1\mathbf{\tilde{A}}_{I_{k+1}})^T \\
\hat{\mathbf{x}}_{k+1} & = \mathbf{\tilde{A}}_{I_{k+1}} \mathbf{\gamma}_{k+1}
\end{align}
\item $k \leftarrow k + 1$; Repeat steps 2-4 $T_0$ times.
\end{enumerate} 

\textbf{Dictionary update}
The dictionaries $\mathbf{A}^H_i$ and $\mathbf{A}^L_i$ can now be obtained as:
$$\mathbf{A}^H_i = \mathbf{\Gamma}_i^T(\mathbf{\Gamma}_i\mathbf{\Gamma}_i^T)^{-1}$$ \\
$$\mathbf{A}^L_i = \mathbf{\Gamma}_i^T(\mathbf{\Gamma}_i\mathbf{\Gamma}_i^T)^{-1}.$$
Further the dictionary atoms are normalized to unit norm in feature space:
$$\mathbf{A}^H_{i,j} = \frac{\mathbf{A}^H_{i,j}}{\sqrt{(\mathbf{A}^H_{i,j})^T \boldsymbol{\mathcal{K}}^H(\mathbf{X}^H_{i},\mathbf{X}^H_{i})\mathbf{A}^H_{i,j}}}, \; j = 1, \cdots, K$$   
$$\mathbf{A}^L_{i,j} = \frac{\mathbf{A}^L_{i,j}}{\sqrt{(\mathbf{A}^L_{i,j})^T \boldsymbol{\mathcal{K}}^L(\mathbf{X}^L_{i},\mathbf{X}^L_{i})\mathbf{A}_{i,j}}}, \; j = 1, \cdots, K$$
\end{document}